\documentclass{article}
\pdfoutput = 1

\usepackage[nonatbib, preprint]{neurips_2020}

\usepackage[utf8]{inputenc} % allow utf-8 input
\usepackage[T1]{fontenc}    % use 8-bit T1 fonts
\usepackage{hyperref}       % hyperlinks
\usepackage{url}            % simple URL typesetting
\usepackage{booktabs}       % professional-quality tables
\usepackage{amsfonts}       % blackboard math symbols
\usepackage{nicefrac}       % compact symbols for 1/2, etc.
\usepackage{microtype}      % microtypography
\usepackage{bm}
\usepackage{enumitem}
\usepackage{graphicx}
\usepackage{algorithm}
\usepackage{algorithmic}
\usepackage{amsmath}
\usepackage{subfigure}
\usepackage{authblk}
\usepackage[toc, page]{appendix}

\newcommand{\ie}{\emph{i.e., }}
\newcommand{\eg}{\emph{e.g., }}

\newcommand{\etc}{\emph{etc.}}

\title{Data Augmentation View on Graph Convolutional Network and the Proposal of Monte Carlo Graph Learning}

\author[1]{Hande Dong}
\author[2]{Zhaolin Ding}
\author[1]{Xiangnan He}
\author[3]{Fuli Feng}
\author[1]{Shuxian Bi}

\affil[1]{University of Science and Technology of China, Hefei, China \authorcr \{donghd@mail.ustc.edu.cn, hexn@ustc.edu.cn, stanbi@mail.ustc.edu.cn\}}
\affil[2]{North Carolina University \{zding8@ncsu.edu\}}
\affil[3]{National University of Singapore, Singapore \{fulifeng93@gmail.com\}}

\begin{document}

\maketitle
\begin{abstract}
    Today, there are two major understandings for graph convolutional networks, \ie in the spectral and spatial domain.
    But both lack transparency.
    In this work, we introduce a new understanding for it -- data augmentation, which is more transparent than the previous understandings.
    Inspired by it, we propose a new graph learning paradigm -- Monte Carlo Graph Learning (MCGL).
    The core idea of MCGL contains: 
    (1) Data augmentation: propagate the labels of the training set through the graph structure and expand the training set; 
    (2) Model training: use the expanded training set to train traditional classifiers.
    We use synthetic datasets to compare the strengths of MCGL and graph convolutional operation on clean graphs.
    In addition, we show that MCGL's tolerance to graph structure noise is weaker than GCN on noisy graphs (four real-world datasets). 
    Moreover, inspired by MCGL, we re-analyze the reasons why the performance of GCN becomes worse when deepened too much: 
    rather than the mainstream view of over-smoothing, we argue that the main reason is the graph structure noise, and experimentally verify our view. 
    The code is available at \url{https://github.com/DongHande/MCGL}.
    
\end{abstract}
\section{Introduction}

Graph data is common in real-world applications. 
Many machine learning methods have been designed for graph, which are collectively referred to as graph learning~\cite{zhu2003semi,zhou2004learning,talukdar2009new,belkin2006manifold,Perozzi2014,Ma2019}.
The specific task in this paper is a graph-based semi-supervised learning: 
Given graph structure, features of nodes in graph, the label of a subset of nodes; 
the goal is to predict for the remaining nodes in the graph. 
An important assumption in some graph learning algorithms is \textit{local homogeneity}~\cite{DBLP:conf/icml/YangCS16},
\ie connected nodes tend to be similar and with the same label.
In recent years, many researches on graph learning focus on Graph Convolutional Network (GCN)~\cite{kipf2017gcn,DBLP:conf/iclr/VelickovicCCRLB18,DBLP:conf/iclr/XuHLJ19,klicpera_predict_2019}.
The formula of a one-layer GCN is: $\bm H^{(k+1)}=\sigma (\hat {\bm A } \bm H^{(k)} \bm W^{(k)})$,
where $\bm H^{(k)}$ denotes the $k$-th layer representation of all nodes, $\hat {\bm A }$ the normalized adjacency matrix (\eg $\hat {\bm A }$ = ${\bm D^{(-1/2)} } {(\bm A + \bm I)}\bm D^{(-1/2)} $, $\bm D$ the degree matrix of $\bm A + \bm I$), $\bm W^{(k)}$ the parameter matrix, and $\sigma (\cdot)$ the non-linear function. 
Based on the above formula, we can separate GCN as two parts: \textit{graph convolutional operation (GCO)}, \ie $\bm P^{(k)} = \hat {\bm A } \bm H^{(k)}$, and a traditional neural network (NN), \ie $\bm H^{(k+1)}=\sigma (\bm P^{(k)} \bm W^{(k)})$.
The interpretability of NN is a traditional problem.
This work mainly tries to interpret GCO.

%Using structure information, GCO aggregates information from neighboring nodes to the central node;
%using feature information, GCO then transforms features into the label space;
%finally GCN trains the labeled nodes in neural network.

Today, there are two understandings on GCO: spectral domain~\cite{kipf2017gcn} and spatial domain~\cite{DBLP:conf/iclr/XuHLJ19, klicpera_predict_2019}. 
The understanding in spectral domain interprets the model by graph signal processing: representation of the nodes is regarded as the graph signal, and processed by the convolution method on the graph.
GCO is the low frequency approximation of graph signal. 
The understanding in spatial domain interprets the model by information propagation: representation of nodes in the graph is regarded as the information of the nodes, and GCO is message-passing and receiving.
GCO makes the nodes get better representations by receiving the information of the neighbor nodes. 
%In this way, GCO is a kind of representation learning. 

\begin{figure}[t]
    \centering
    \vspace{-0.2cm}
    \includegraphics[width=\textwidth]{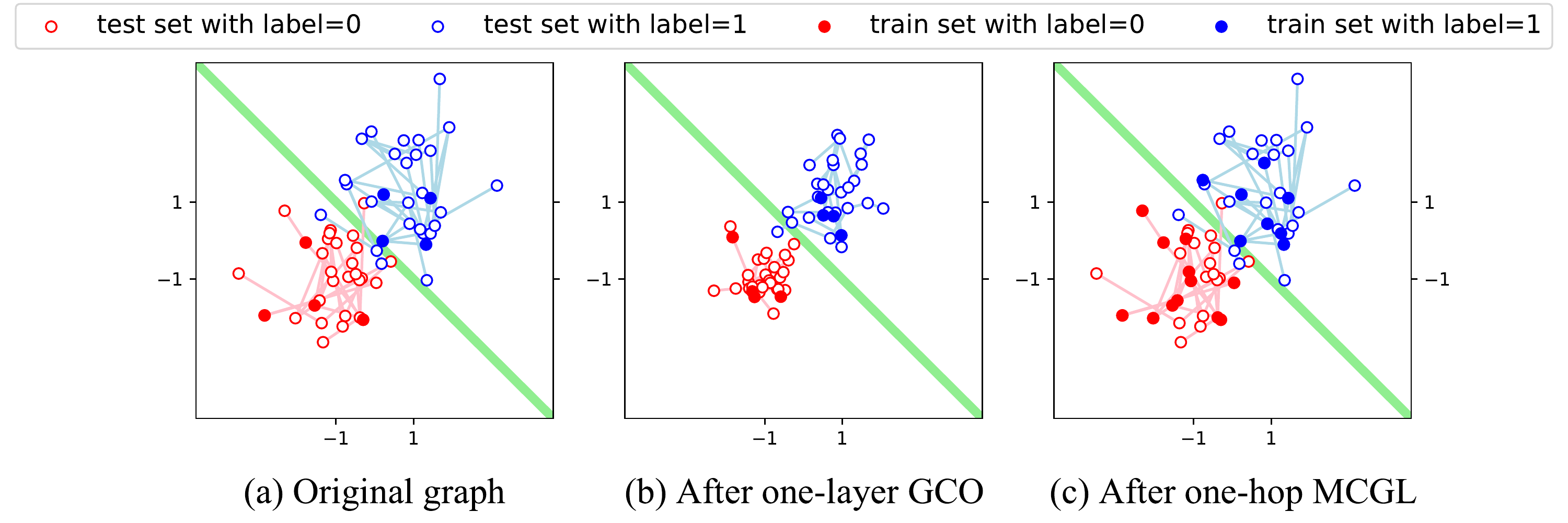}
    \vspace{-0.7cm}
    \caption{Data augmentation based on graph structure. (a) original graph; (b) after one-layer GCO, the intra-class variance is smaller, making the inter-class boundary more obvious; (c) after one-hop MCGL, each class has more training samples, making it easier to learn a good classification boundary. }
    %\caption{Data augmentation based on graph structure. (a). the original features of two groups of Gaussian distributed \textit{i.i.d} nodes. (b) and (c) show the transformed data after one-layer GCO and one-hop MCGL.}
    \label{model_comparison}
    \vspace{-0.2cm}
\end{figure}

But both perspectives lack transparency.
The view in spectral domains is abstract, and it is hard to understand the meaning of graph signal in spectral space for many people.
The view in spatial domain is only a qualitative analysis, and can't explain why and when aggregating neighbors leads to better representation.
In this work, we introduce a new understanding of GCO -- \textbf{data augmentation}.
GCO makes the connecting nodes closer.
Under the assumption of local homogeneity, the representations of nodes in same class get closer in representation space, which makes the classification boundaries more obvious.
Thus, GCO can be regarded as data augmentation.
As Figure~\ref{model_comparison}.(a) and (b) shows, after one-layer GCO, the variance of the features of the nodes in each class becomes smaller while the average value almost remains unchanged.
Machine learning in (b) is easier than (a).

%, the nodes with same label tend to aggregate to the central region of themselves, because most of their neighbors have the same label.
%By separating nodes with different labels, graph convolutional operation makes classification easier.

%The Coefficient of variation among features of each class of nodes  
%$c_v = \frac{\sigma}{\mu}$ decreases, separating different %classes of nodes in both training set and test set efficiently.
%Therefore, GCO implements data augmentation from the level of data representation.

GCO can be regarded as a kind of data argumentation in representation space.
Inspired by this, we propose a new paradigm to implement data argumentation in label space - \textbf{Monte Carlo Graph Learning} (MCGL). 
%From the same perspective, we propose a new graph learning paradigm that implements data augmentation - \textbf{Monte Carlo Graph Learning} (MCGL).
As Figure \ref{model_comparison}.(c) shows, the representations of all nodes remain unchanged, 
MCGL instead delivers labels of nodes in the training set to their neighbors and assign them as pseudo-labels to expand the training set.
MCGL follows three steps: 
(1) Graph Monte Carlo sampling: take the nodes in the training set as root nodes, perform $K$ times MC sampling according to graph structure; 
%Monte Carlo sampling over neighbors of each node according to the graph structure.
(2) pseudo-labels: assign the labels of root nodes to the sampled neighbor nodes;
(3) Learning: train the base machine learning model (such as MLP, LR, SVM) with the sampled nodes. 
$K$ in step (1) represents the \textit{depth} of MCGL.

%1) perform Monte Carlo (MC) sampling over neighbors of each labeled node, assign the neighbor nodes the same label (\ie the pseudo-label);
%2) and combine the original data with pseudo-labeled data as augmented data to train standard classification.

The assumption of local homogeneity in graph structure provides reasonableness for MCGL. 
The edges on a graph can be divided into two categories: \textit{good edges} and \textit{bad edges}. 
Good edges refer to those connect two nodes with the same label, and bad edges are the opposite. 
If all edges in the graph are good edges (completely satisfying the assumption of local homogeneity), we call it a \textit{clean graph}. 
Otherwise, we call it a \textit{noisy graph}. 
By experiments on clean graphs, we compare the respective strengths of GCO and MCGL in a visual way:  
MCGL is good at dealing with datasets with a non-linear boundary or community characteristics, while GCO is good at datasets of large variance.
By experiments on noisy graph, we show that MCGL has a lower tolerance to the noise than GCO.
% By experiments on noisy graph, we sho w that MCGL's tolerance to noise is weaker than GCN on noisy graph. 

Moreover, our theoretical analysis draws a conclusion that: MCGL cannot go too deep because the bad edges cause too many samples incorrectly labeled.
Inspired by this, we re-analyze the reason why the performance of GCN becomes worse with the depth increasing: 
due to the presence of graph structure noise, more and more nodes with different labels are aggregated, which makes it more difficult to distinguish the nodes. 
We point out that over-smoothing should be the phenomenon rather than the cause that limits the performance of deep GCN.

\section{Comparison on clean graphs}

\begin{figure}[t]
\centering 
\subfigure[Traditional XOR problem] %第一张子图
{
\centering          %子图居中
\includegraphics[scale=0.5]{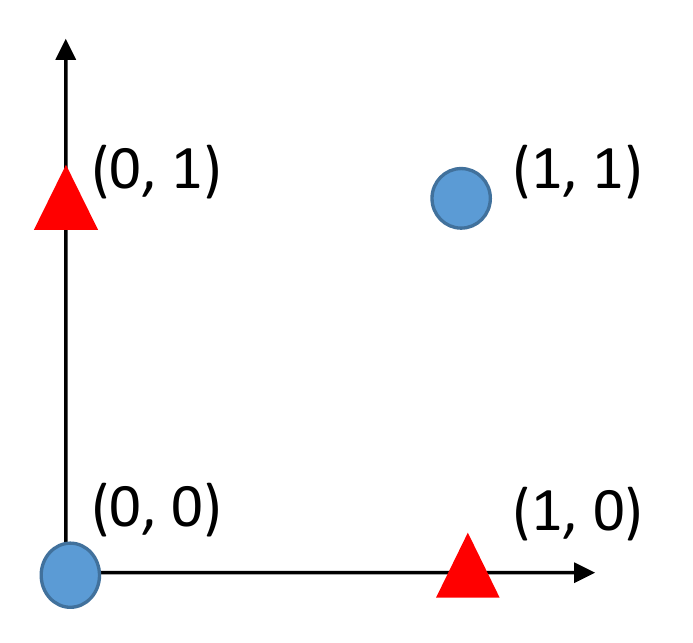}   %以pic.jpg的0.5倍大小输出
}
\subfigure[Graph-XOR problem] %第二张子图
{
\centering      %子图居中
\includegraphics[scale=0.5]{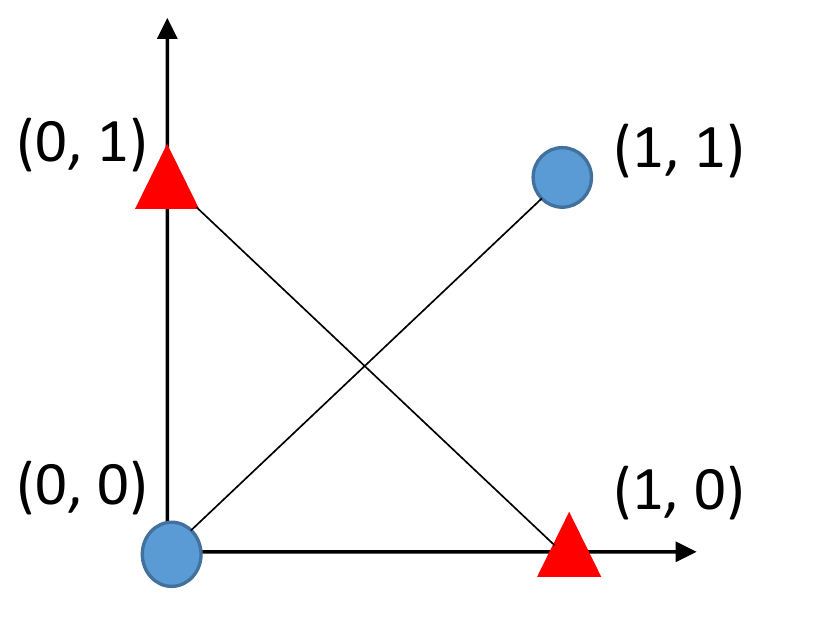}   %以pic.jpg的0.5倍大小输出
}
\vspace{-0.2cm}
\caption{Two XOR problems: compared to (a), (b) uses the samples as nodes in graph, and connects the nodes with the same labels}  %大图名称
\vspace{-0.4cm}
\label{XOR}
\end{figure}

Adding two edges to traditional XOR problem, Figure \ref{XOR}.(b) shows Graph-XOR problem.
Although the XOR graph is free of noise, GCO fails to deal with it.
After one-layer GCO, four nodes are aggregated to the same position $(0.5, 0.5)$,
and no further operation can separate them again.
There will be no way to distinguish them and make a classification.
Meanwhile, MCGL is able to deal with Graph-XOR problem as long as the base model has non-linear classification capability (such as MLP).

%MCGL is better than GCN in graph XOR case. Is there some cases where GCO is better than MCGL? Or what are the strengths and weaknesses of the two models? 
%In this section, we will compare MCGL and GCN through several synthetic datasets, to find their perspective advantages. 
%%to find out which one works better with specific type of datasets.
%The graph are all clean in this section, \ie connection nodes all have the same labels. 
%This simplification makes those graphs suitable for graph learning.

In this section, we compare GCO and MCGL by experiments on synthetic clean graphs in a visual way.
Every dataset consists of 60 nodes in two classes, identified as red and blue points in the figures.
The solid points make up the training set.
The feature is the two-dimensional coordinates of the points.
The edges (all are good edges), represent the graph structure.
%The cleanness of graphs make them suitable for graph learning.
We will show that GCO is good at dealing with datasets of large variance, and MCGL is good at datasets with a non-linear boundary or community characteristics.

\begin{figure}[t]
    \centering 
    \vspace{0cm}
    \includegraphics[width=\textwidth]{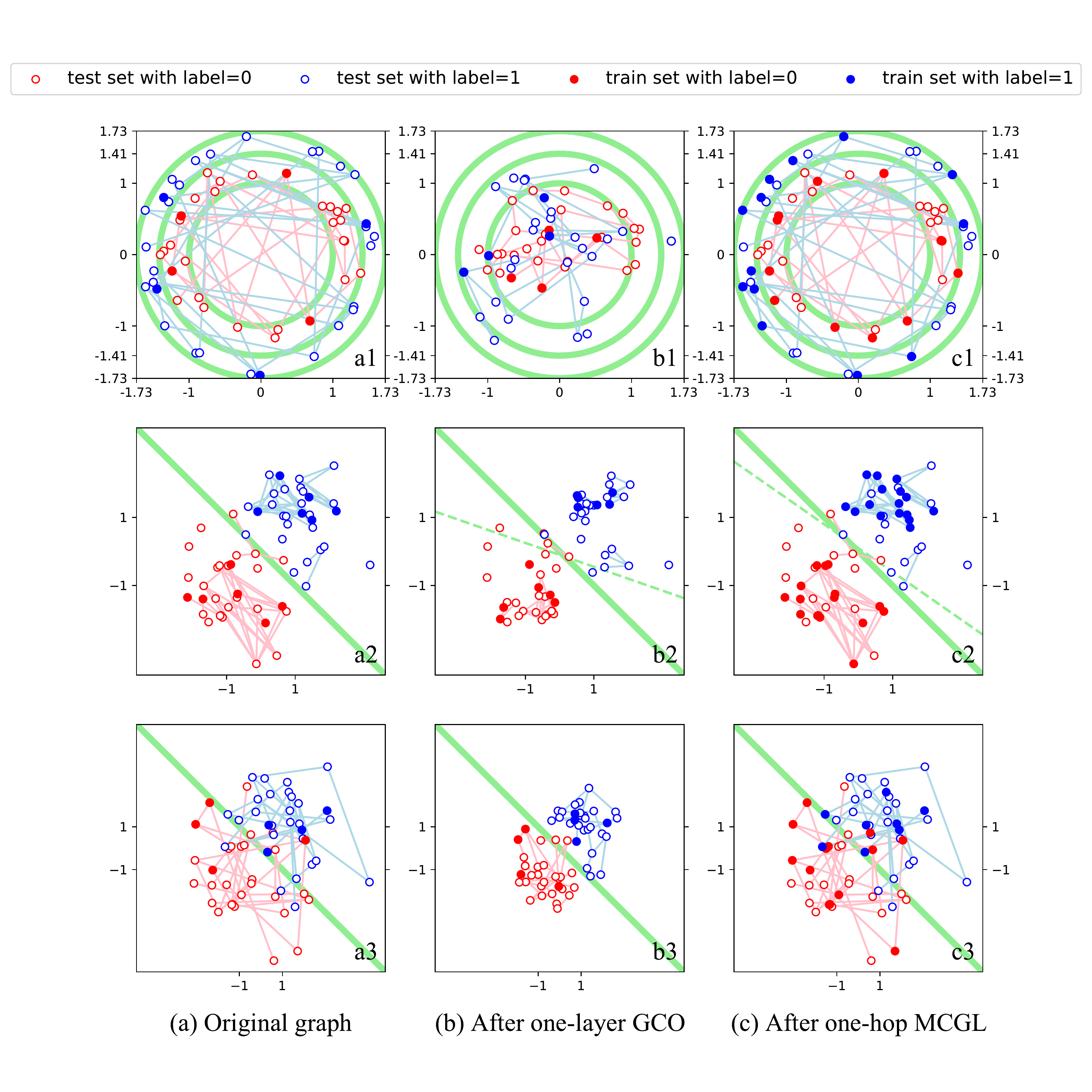}
    \vspace{-1.2cm}
    \caption{The original graph and the transformed graphs after one-layer GCO and one-hop MCGL. Row 1: non-linear boundary; Row 2: community characteristics; Row 3: large variance.}
    \vspace{-0.4cm}
    \label{comp}
\end{figure}

\paragraph{Non-linear boundary}
Since GCO aggregates the representations of connected nodes to the average value of them,
it leads to intersection of different classes if the boundary is non-linear.
Thus, GCO is not good at dealing with non-linear problems.
Taking concentric circles as an example: 
Two groups of points are uniformly distributed in two donut-like regions with the same area, as in Figure \ref{comp}.(a1).
After one-layer GCO, red points and blue points tend to be aggregated to the same position -- near the center of the circles, as in Figure \ref{comp}.(b1), and the inter-class boundary becomes vague.
So the prediction accuracy will be low.
% And it becomes even worse as GCO gets deeper.
Meanwhile, since MCGL implements data augmentation in label space rather than representation, there is no impact on MCGL whether the boundary is linear or non-linear.
As in Figure \ref{comp}.(c1), after one-hop MCGL, the training set is expanded, thus can reveal more details about the boundary, helping increase the prediction accuracy.

% \begin{figure}[h]
%     \centering 
%     \vspace{0cm}
%     \includegraphics[width=\textwidth]{comp_non-linear.pdf}
%     \vspace{-0.5cm}
%     \caption{Two groups of points uniformly distributed in concentric circles, and the transformed data after one-layer GCO and one-hop MCGL.}
%     \vspace{0cm}
%     \label{non-linear}
% \end{figure}

\paragraph{Community characteristics}
Community characteristics mean that the nodes can be divided into multiple groups (communities) with no (or few) interconnections between them.
GCO only aggregates the nodes within a group toward their center, so when the training set is unevenly distributed in each group, GCO cannot handle it well. 
As a simple example in this case, two classes of Gaussian distributed \textit{i.i.d} points with the variance of 1 are centered at $(-1, -1)$ and $(1, 1)$ respectively, as in Figure \ref{comp}.(a2).
In each class, there are a major community containing most of the points, and several small communities surrounded.
The training set is mainly distributed in both major communities.
As Figure \ref{comp}.(b2) shows, after one-layer GCO, the boundary according to the training set (dotted line) is far from the real boundary (solid line).
This is because new representations are aggregated to respective group centers.
It magnifies the unevenness of the training set, \ie the distribution of the training set in representation space becomes more uneven. 
Meanwhile, as in Figure \ref{comp}.(c2), MCGL gives a more accurate boundary than GCO.
It is because the training set is expanded with nodes distributed more widely after one-hop MCGL.

% \begin{figure}[h]
%     \centering 
%     \vspace{0cm}
%     \includegraphics[width=\textwidth]{comp_community.pdf}
%     \vspace{-0.5cm}
%     \caption{Two groups of Gaussian distributed \textit{i.i.d} points with community characteristics, and the transformed data after one-layer GCO and one-hop MCGL.}
%     \vspace{0cm}
%     \label{community}
% \end{figure}

\paragraph{Large variance}
When the variance of nodes in the same class is large, GCO becomes suitable.
GCO can aggregate connected nodes and reduce the variance from too large to a more proper level.
MCGL instead implements data augmentation in label space, while the large variance remains unchanged.
Take the two groups of Gaussian distributed \textit{i.i.d} points as an example again.
In this case, the variance among either group is set as 2, so some points are intersected with the other class in the original graph, as in Figure \ref{comp}.(a3).
So the training set hardly reveals an accurate boundary. 
However, after one-layer GCO, as the variance becomes smaller, the classification boundary becomes obvious as in Figure \ref{comp}.(b3).
GCO does well in this case.
Meanwhile, MCGL assigns many pseudo-labels, as in Figure \ref{comp}.(c3).
The expansion of training set may be helpful for learning, however, MCGL does not perform as well as GCO. 

%GCO is shown to be more suitable in this case.

% \begin{figure}[t]
%     \centering 
%     \vspace{0cm}
%     \includegraphics[width=\textwidth]{comp_large_variance.pdf}
%     \vspace{-0.5cm}
%     \caption{Two groups of Gaussian distributed \textit{i.i.d} points with larger variance, and the transformed data after one-layer GCO and one-hop MCGL.}
%     \vspace{0cm}
%     \label{large_variance}
% \end{figure}
\section{Comparison on noisy graphs}

In last section, we have compared MCGL and GCO on clean graphs.
However, graphs collected from the real world always contain graph structure noise.
Here, we will make a comparison between MCGL and GCO on noisy graphs.
The graph structure \textit{noise rate}, \ie the ratio of bad edges to all edges, can measure the strength of graph structure noise.

\begin{figure}[t]
    \centering 
    \vspace{0cm}
    \includegraphics[width=\textwidth]{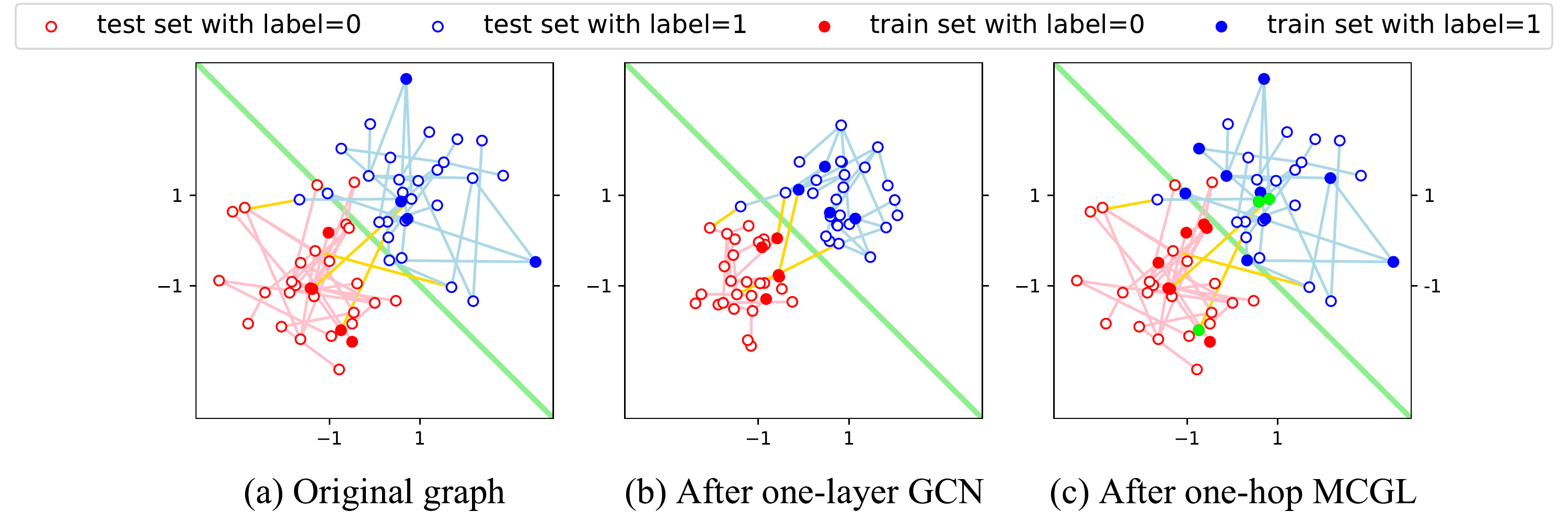}
    \vspace{-0.6cm}
    \caption{Two groups of Gaussian distributed \textit{i.i.d} points with the presence of graph structure noise, and the transformed data after one-layer GCO and one-hop MCGL. Green points in (c) are those sampled and get incorrect pseudo-labels.}
    \vspace{0cm}
    \label{comp_noise}
\end{figure}

Graph structure noise has a direct impact on MCGL. It leads to incorrect pseudo-labels and affects MCGL in a hard way.
On the contrary, graph noise has an indirect impact on GCO, because it only causes GCO to aggregate the representations of nodes with different labels as a part. 
Graph structure noise affects GCO in a relatively soft way.
Therefore, we point out that MCGL have a lower tolerance to the noise than GCO, \ie when the noise is large, GCO performs better; otherwise, MCGL performs better.
Figure \ref{comp_noise} illustrates the different processing methods on noisy graphs by GCO and MCGL.
Next, we will verify our view using four real-world datasets -- CORA~\cite{mccallum2000automating}, CiteSeer~\cite{sen2008collective}, PubMed~\cite{namata2012query}, and MS Academic~\cite{shchur2018pitfalls}.

\paragraph{Model}
We introduce a simple implementation of MCGL -- MCGL-UM model (MCGL-uniform-MLP).
In MCGL-UM, Monte Carlo sampling follows the uniform distribution, and the base model is MLP.
The training process is shown in Algorithm~\ref{MC_sample_graph_algorithm_1}.
With the well-trained model, prediction for all nodes in the test set can be made in the inference process, \ie $\bm y_i^{(0)}=f_\theta(\bm x_i)$.
%However, directly using $\bm y=f_\theta(\bm x)$ is not a good choice because it does not utilize the \textit{local homogeneity} of the graph structure.
Moreover, we recursively aggregate the predictions of neighbor nodes to improve accuracy, 
%As an improvement, we recursively aggregate neighbors $K$ times to improve performance,
\ie $\bm y_i^{(k+1)} = \sum_j p_j^{(i)}\bm y_j^{(k)},\, j \in \mathcal{N}_i$.
The iterative form is that $\bm Y^{(k+1)}=\hat{\bm A}\bm Y^{(k)}$, and $\bm Y^{(0)}=f_\theta(\bm X)$.
The sub-graphs of $K$-hop neighbors of a node make up a $K$-layer tree, from where we can get a more thorough understanding on MCGL.
%In Monte Carlo sampling tree (Figure~\ref{mc_sample}(b)), 
During the training process, a leaf node is sampled from the root node of the tree in a top-down manner.
The model is trained using the leaf nodes.
During the inference process,
all leaf nodes of the tree are predicted using the trained model, 
then all the predictions are aggregated to the root node of the tree in a bottom-up manner to predict the root node.
%\begin{figure}[t]
%    \centering 
%    \includegraphics[width=0.3\textwidth]{mc_tree.pdf}
%    %\vspace{-0.2cm}
%    \caption{Monte Carlo sampling tree in MCGL: the root is a node in the training set and the child nodes are the neighbors of the parent node. Each time we perform Monte Carlo sampling, the sampling tree is deepened by one layer.}
    %\vspace{-0.4cm}
%    \label{mc_sample}
%\end{figure}

\begin{algorithm}[t]
\caption{Training procession of MCGL-UM}
\label{MC_sample_graph_algorithm_1}
\textbf{Input}: graph structure $G = (V, E)$; feature matrix $\bm X$; 
label matrix $\bm Y$; training set id $T$; model $y=f_\theta(\bm x)=MLP(\bm x)$\\
\textbf{Parameter}: parameters in model $f_\theta(\bm x)$\\
\textbf{Hyper-parameter}: Monte Carlo sampling degree $K$; batch size $N$;
other hyper-parameter for model $f_\theta(\bm x)$ (learning rate, l$_2$-norm, \etc)\\
\textbf{Output}: the well trained model $y=f_\theta(\bm x)$\\
\begin{algorithmic}[1] %[1] enables line numbers
\STATE Calculate probability distribution $P^{(i)} (x=j) = p^{(i)}_{j} = 1/|\mathcal{N}_i|,j \in \mathcal{N}_i$ for all nodes according to graph structure $G=G(V,E)$.
\STATE Initialize model $y=f_\theta(\bm x)$.
\WHILE{ $\,$model $f_\theta(\cdot)$ doesn't converge$\,$ }
\STATE batch list $L=\{\,\}$
\WHILE{$n$ from 1 to $N$}
\STATE sample an id $i^n$ from $T$    \quad\quad\quad \#\textit{the root of MC sampling tree}
\STATE $i^n{(0)}=i^n$
\WHILE{$k$ from 1 to $K$}
\STATE $i^n(k) = MC\_Sampling(i^n(k-1))\,$   \quad\quad\quad \#\textit{the sampling tree gets deeper by one layer}
\ENDWHILE
\STATE add $(\bm x_{i(K)}^n,y^n_i)$ to batch list $L$
\ENDWHILE
%\STATE use $N$-degree graph Monte Carlo sampling to sample a batch data: 
%$\{(\bm x^k_{i(n)}, \bm y^k_i)$, where $k=1,2,\cdots, K\}$
\STATE train $f_\theta(\bm x)$ with batch data $L$ 
\ENDWHILE
\STATE \textbf{return} well-trained model $y=f_\theta(\bm x)$
\end{algorithmic}
\end{algorithm}

\paragraph{Experiment setup} The original noise rate of CORA, CiteSeer, PubMed and MS Academic are $19.00\%, 26.45\%, 19.76\%, 19.19\%$ respectively.
We manually reduce the noise rate on each graph to different levels, train MCGL-UM and traditional GCN model on modified graphs, and compare their accuracy under different noise rate.
The depth of GCN model and both sampling and inference processes of MCGL-UM are set as 2.

\paragraph{Dataset}
CORA, CiteSeer and PubMed are citation graphs, where a node represents a paper, and an edge between two nodes represents that the two papers have a citation relationship.
MS Academic is co-author graph, an edge in the graph represents the co-authorship between two papers.
Table \ref{dataset} shows the statistics and the data split of the datasets.

\begin{table}[t]
\centering
\caption{Dataset statistics}
    \begin{tabular}{ccccccc}
    \hline
    Dataset   & Type      & Nodes  & Edges   & Features & Classes & Training/Validation/Test  \\
    \hline
    CORA        & Citation  & 2,708  & 5,427   & 1,433    & 7   & 140/500/1,000 \\
    CiteSeer    & Citation  & 3,327  & 4,732   & 3,703    & 6   & 120/500/1,000 \\
    PubMed      & Citation  & 19,717 & 44,338  & 500      & 3   & 60/500/1,000 \\
    MS Academic & Co-author & 18,333 & 163,788 & 6,805    & 15  & 300/500/17,533 \\
    \hline
    \end{tabular}
\label{dataset}
\end{table}

\begin{figure}[t]
    \centering 
    \vspace{0cm}
    \includegraphics[width=\textwidth]{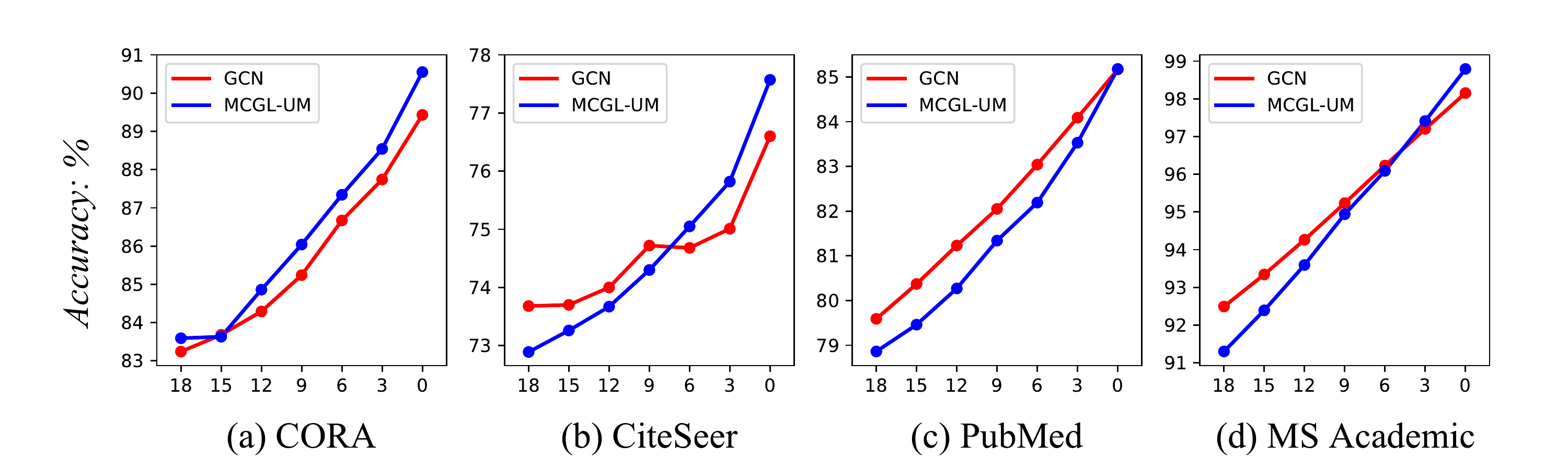}
    \vspace{-0.5cm}
    \caption{The trend of accuracy of MCGL-UM and traditional GCN with decreasing noise rate (\%).}
    \vspace{0cm}
    \label{acc_to_noise}
\end{figure}

\paragraph{Experimental results}
Figure \ref{acc_to_noise} shows the trend of accuracy of MCGL-UM and traditional GCN with decreasing noise rate.
By removing bad edges according to the ground truth of the labels of all nodes, we can reduce the noise rate of graphs.
The x-axis represents the percentile noise rate, and the y-axis represents the percentile accuracy.
We can find that: 
(1) Although the accuracy of MCGL-UM is always slightly better than GCN on CORA and the opposite on PubMed, the improvement of MCGL-UM is always greater than GCN as noise rate decreases.
(2) The overall trend on four datasets all follow that when the noise rate is relatively high, GCN is slightly better than MCGL-UM; otherwise, MCGL-UM is slightly better than GCN. 
Therefore, we can conclude that MCGL-UM is more vulnerable to graph structure noise than GCN, but it performs better on clean graphs.

% 中心论点：MCGL对噪声的容忍能力要弱于GCO--即噪声小的时候，MCGL好；噪声大的时候，GCO好。

% 论证过程：
% 第一步：说明现实世界中的图并不总是clean的，the graph structure collected from the real world will always have some noise.
% the noise rate means... which can measures by the rate of bad edges among all edges.
% 当有噪声的时候，MCGL和GCO各自怎么处理，MCGL：。。。。。。
% 图说明我们的看法（有噪声）。

% 我们将会在真实数据集下验证我们的想法。
% 因为MCGL只是一个paradigm，而不是具体的模型，因此我们要对MCGL设计出具体的模型，在这里，我们用一个简单的MCGL的实现：MCGL-UM。
% 下面，我们首先介绍MCGL-UM；最后在真实数据集下的实验介绍在数据集下MCGL-UM和GCN的对比，展示了MCGL对噪声的容忍能力更差。

% \subsection{Model}---MCGL-UM

% \subsection{Experiment}

% \paragraph{Dataset}

% \paragraph{Expe result}

% Conclusion
\section{Why can't deep?}

As many previous works observed, when the GCN model goes too deep, its accuracy will drop, and the best depth ($K$) is always 2.
The mainstream view of what restricts the depth is \textit{over-smoothing},
% As $K$ grows, the representations of nodes ($\bm H^{(K)}$) get closer.
% When $K$ approaches infinity, the representations tend to be in a stable state independent of the origin ($\bm H^{(0)}$).
\ie when the $K$ reaches a threshold, $\bm H^{(K)}$ will become too close to be distinguishable, where $\bm H^{(K)} = \hat {\bm A}^K\bm H^{(0)}$ is the representation of nodes in matrix form.
% It seems plausible from the perspective of element.
% The $K$-layer GCO is described in the form of element as $\bm h_p^{(K)} = \sum_{j \in \mathcal{N}_p^{(K)}}\alpha_j^{(p)} \bm h^{(0)}_j$, for node $p$.
% For a pair of close nodes (say, node $p$ and $q$), their $K$-hop neighbors have intersections.
% $\bm h_p^{(K)}$ and $\bm h_q^{(K)}$ have some terms in common, making them correlated with each other.
% As $K$ increases, the proportion of same terms increases as well, making nodes indistinguishable.
% However, when it comes to clean graphs, deep GCN can work perfectly (see Figure \ref{noise}.(d) first row).
% Instead, we point out that it is the graph structure noise that limits the depth of graph learning models.
However, we disagree with this view and argue that the main factor limiting the depth of GCO is graph structure noise.

%We first find that the depth of MCGL is limited by the presence of graph structure noise.
In MCGL, we have 100\% confidence in the labels of the training set when $K = 0$; 
after taking a step ($K = 1$), with the expansion of the training set, some nodes will get incorrect pseudo-labels, which decreases the confidence in the labels; 
and as $K$ continues to grow, training set expands bigger, but the confidence of the training sets will further decrease.
Therefore, large $K$ on one hand can augment data, which is beneficial to learning;
on the other hand causes the noise data, which is harmful to learning.
The best $K$ can be seen as a result of trade-off between the two sides.
When $K$ is just beginning to grow, the benefit of data augmentation is greater; and when $K$ continues to grow, the noise does more harm. 
This explains why the depth of MCGL can't be too deep. 

\begin{figure}[t]
    \centering 
    \vspace{-0.1cm}
    \includegraphics[width=\textwidth]{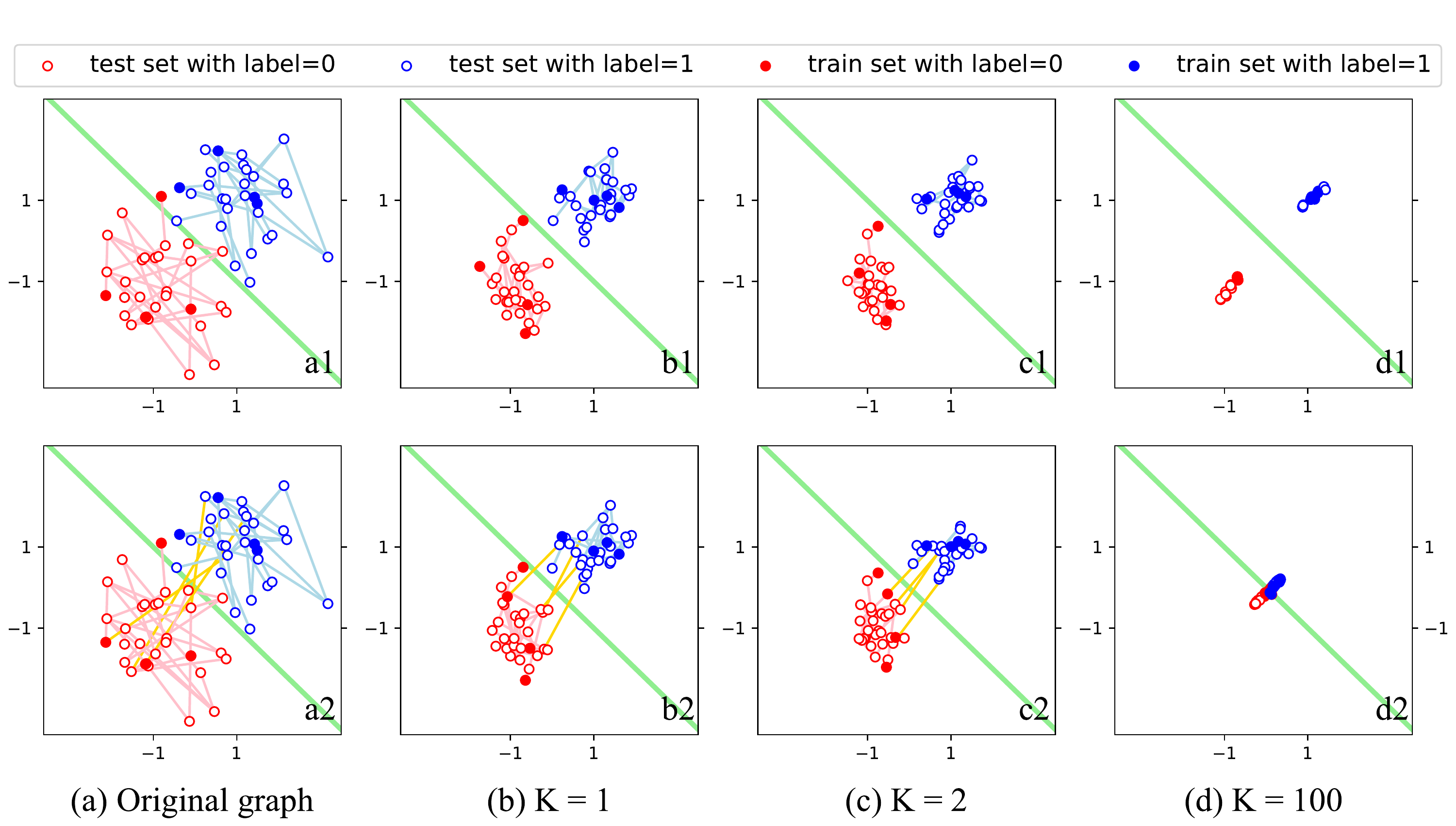}
    \vspace{-0.6cm}
    \caption{Deep GCO in clean graph and noisy graph. Row 1: clean graph; Row 2: noisy graph.}
    \vspace{-0.2cm}
    \label{noise}
\end{figure}

Inspired by this, we propose that the reason for limiting the depth of GCO is also graph structure noise. 
As the depth of GCO deepens, more and more nodes of different classes will be aggregated, which is harmful.
Figure~\ref{noise} is a visual display of this view. 
The first line shows the impact of GCO on clean graph; the second line shows the impact of GCO on noisy graph. 
On clean graph, as GCO deepens the intra-class variances of nodes reduce, but the mean of each class remains basically unchanged.
Therefore, the deeper the GCO, the easier to distinguish the classes (d1).
However, on noisy graph, with the reduction of intra-class variance, the means of different classes begin to approach due to the existence of bad edges. 
Therefore, too deep GCO will make it difficult to distinguish nodes of different classes (d2). 
On clean graph, the depth GCO causes intra-class over-smoothing, but this over-smoothing doesn't adversely affect the final classification effect; 
However, on noisy graph, the depth GCO causes inter-class over-smoothing, which directly causes classification result declining. 
Therefore, it is unreasonable to think that over-smoothing is the main reason for limiting the depth of GCO. 
In fact, over smoothing is a phenomenon of deep GCO, and the main reason for limiting the depth of GCO is graph structure noise.

%Graph structure noise has a major impact on deep GCO as well.
%Take two groups of Gaussian distributed \textit{i.i.d} points as an example again.
%Figure \ref{noise}.(a1) shows an original "clean" graph without any noise in its structure.
%Figure \ref{noise}.(a2) shows an original "noisy" graph with a few bad edges (yellow lines) in it.
%When GCO is shallow ($K = 1$ and $2$), the difference between transformed clean graph and noisy graph is little, as in Figure \ref{noise}.(b) and (c).
%However, weak noise causes a significant effect on 100-layer GCO (see Figure \ref{noise}.(d)).
%On the clean graph, deep GCO greatly reduces the intra-class variance of nodes, so that both groups are tightly aggregated.
%Thus, a boundary can be decided much more flexible to make an accurate classification.
%But on the noisy graph, not only the intra-class variance is reduced, the inter-class variance of nodes is reduced as well.
%The transformed representations are all aggregated to the Origin Point, making it hard to distinguish nodes in different classes.
%Therefore, we believe that graph structure noise limits the performance of deep GCN.

To further verify our view, we conduct experiments on real-world datasets. 
Instead of the traditional GCN model, we use another GCN model which applies the APPNP strategy \cite{klicpera_predict_2019} (called GCN* in this paper), \ie $\bm Y= softmax (\hat {\bm A }^K MLP(\bm X))$. 
Compared with traditional GCN, GCN* has a constant layer of neural network regardless of the change of the depth of GCO.
It is introduced for the purpose to: (1) prevent gradient vanish and gradient explosion due to deep neural network; (2) prevent overfitting due to too many parameters; 
and most importantly (3) ensure that the comparison between different depth of GCO is fair.

\begin{figure}[t]
    \centering 
    \vspace{-0cm}
    \includegraphics[width=\textwidth]{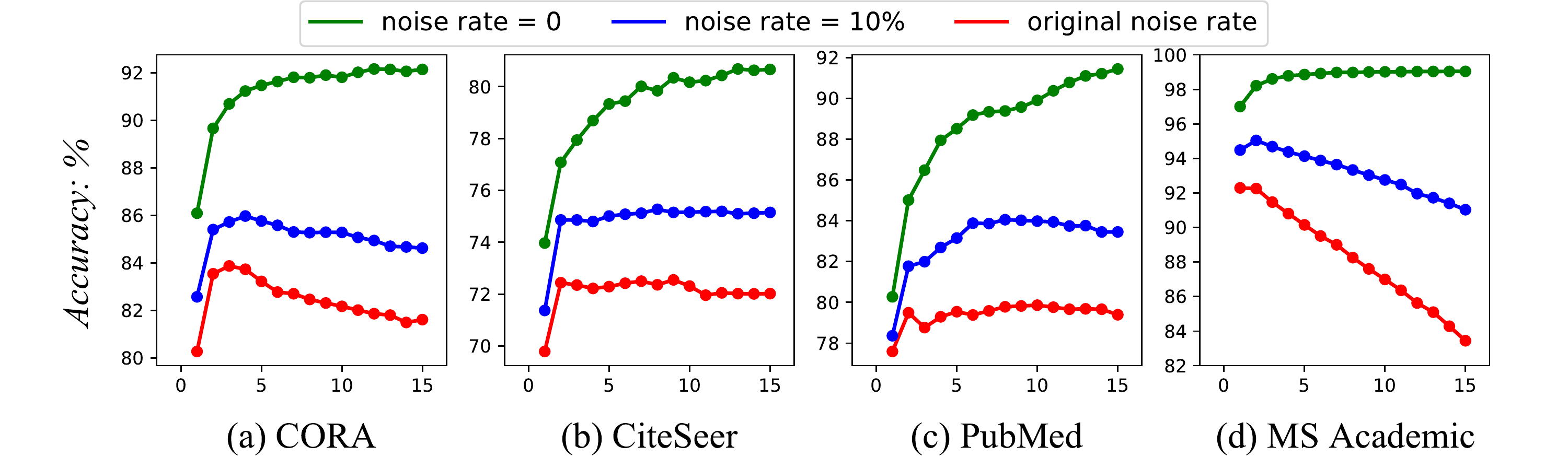}
    \vspace{-0.6cm}
    \caption{The trend of accuracy of GCN* with the depth of GCO under decreasing noise rate (\%).}
    \vspace{-0.4cm}
    \label{acc_to_depth}
\end{figure}

Figure \ref{acc_to_depth} shows the results. 
In the figures, the x-axis represents the depth of GCO (we plot from $K=1$ because the accuracy when $K=0$ are very low and decrease the contrast), and the y-axis represents the percentile accuracy. 
The three lines represent different noise rate after removing bad edges. 
From the figures, we can find that with graph structure noise decreasing, the optimal depth increases. 
When the noise rate becomes 0, the accuracy of GCN* steadily improves til $K=15$. 
By these experiments, we further confirmed our argument about why deep GCO fails.

% From both experiments above, we can conclude that GCN model and MCGL-UM model can achieve better performance with less noise in the graph.
% That is to say, the main factor that limits the depth of graph learning is not over-smoothing, but noise.
% Actually, over-smoothing is also caused by the graph structure noise.
% The main feature of GCO is to aggregate different nodes, which is only based on the information of graph structure.
% The noise affects GCO by leading to incorrect representations.
% Therefore, we conclude that the main factor is graph structure noise, whereas over-smoothing is rather a cause of noise.
% And deeper graph learning can be more accurate by manually reducing the noise.
\section{Related work}

Convolutional neural networks have achieved great success in the field of computer vision~\cite{DBLP:conf/cvpr/HeZRS16}. 
It inspired researchers to apply convolution operation to graph signals~\cite{DBLP:conf/nips/DefferrardBV16, kipf2017gcn}. 
In spectral domain, graph convolutional operations can be understood as a low-pass filter.
It performs low-pass filtering on the information of the graph signal in the spectral domain, and generates an elegant form of GCN, \ie $\bm H^{(k+1)}=\sigma (\hat {\bm A } \bm H^{(k)} \bm W^{(k)})$. 
Both SGCN and APPNP~\cite{DBLP:conf/icml/WuSZFYW19, klicpera_predict_2019} showed that it is not necessary to couple graph convolutional operations and neural network operations together. 
SGCN first performs GCO several times on input features, then uses linear regression model to transform the results into label space. 
Conversely, APPNP first transforms the features into label space using MLP, then performs GCO (generalized GCO -- using graph structure to combine nodes' representations) several times. 
% In spectral domain, GCO can be understood as a low-pass filter, which filters high-frequency signals but leaves low-frequency signals. 
In spatial domain~\cite{DBLP:conf/iclr/XuHLJ19,klicpera_predict_2019}, GCO is understood as an operation to aggregate neighbors, so better representations can be obtained by the aggregation. 

The present work argues that GCO can also be regarded as a method of data augmentation. 
Based on this new cognition, MCGL was proposed. 
During training, MCGL can be regarded as a pattern of label propagation (LP~\cite{zhu2005semi}) then learning parameter. 
During inference, MCGL can be regarded as predicting then GCO. 
MCGL has a close relationship with the traditional LP algorithm. 
Both methods have the process of propagating labels through the graph structure in common; 
the differences between them are that: (1) LP propagates labels for many times until the distribution converges, while MCGL only performs limited times of propagation 
(because when propagation goes too far, the labels will no longer be trustworthy); 
(2) MCGL train parameters in the base model after the propagation of the labels, which is not available in LP. 
There are also some previous works using graph sampling~\cite{NIPS2017_6703} and graph data augmentation~\cite{DBLP:conf/aaai/LiHW18}, but they are still in the framework of GCN, and try to improve GCN. 
MCGL does not use GCN framework in training. 
Besides, \cite{chen2019measuring} demonstrated that graph noise limits the performance of GCN (on 2-layer GCN, the performance becomes better if the graph noise is decreased). 
Further, the present work argues that the graph structure noise is also the reason that limits the depth of GCN. 
In fact, in semi-supervised learning, the pattern of pseudo-labeling then training the model has already been developed by some previous works~\cite{triguero2015self}. 
But those methods generally label the samples based on the distance of their features in Euclidean space.
Different from those methods, MCGL labels the samples based on the graph structure collected from the real world.
\section{Conclusion}

In this work, we introduce a new understanding toward GCO -- GCO can be regarded as a kind of data augmentation.
Inspired by this, we propose a new graph learning paradigm -- Monte Carlo Graph Learning, and a simple implementation of it -- MCGL-UM model.
GCO augments graph data in representation space, whereas MCGL in label space.
Both have strengths and weaknesses. 
Through a series of experiments on synthetic data and real-world datasets, we verified:
(1) On clean graphs, GCO is more suitable at datasets with large variance, whereas MCGL at datasets with a non-linear boundary or community characteristics.
(2) On noisy graphs, MCGL is more vulnerable to the noise than GCO. 

The most important contribution of this paper is to provide a data augmentation understanding toward GCO and MCGL.
MCGL and GCO are two symmetric methods of data augmentation on graphs. 
Both can inspire a deeper understanding of the other model, \eg we analyzed the reasons limiting the depth of MCGL then found that the depth of GCO is limited by the same reasons.
Nevertheless, MCGL and GCO have their own areas of expertise, thus they can complement each other.

\section*{Broader Impact}

From societal perspective, the present work is more of theoretical and does not involve specific ethical issues such as human image synthesis. 
From academic perspective, the focus of this work is on the interpretability of machine learning (specifically, graph convolutional operations).
Interpretability determines whether a model can be trusted and understood by people, which is actually an ethical issue for human-machine relationship.
Only when the two groups in society (humans and machines) can understand (mutually or unidirectionally) can a normal ethical order be constructed.
% Specifically, interpretability allows people to understand why a machine learning model makes mistakes, so that the ethical problems can be solved completely. 
% Further, with understanding the mistakes, the model thus can be corrected.

The interpretability of a model directly determines whether it can be applied in some sensitive areas. 
Take the user profile in the bank system as an example, the relationships among users can construct a graph, which satisfies the local homogeneity assumption, thus is suitable for graph learning.
If one model shows strong effectiveness but lacks interpretability, banks in some countries cannot adopt it.
This is because if the banks use the model to make uninterpretable decisions (such as refusing a user's application for loans without any reason), it is an infringement of user rights and may lead users to sue banks for lacking fairness and transparency.
On the contrary, if the model can be interpreted, banks can find out the reasons when a user complains, then artificially analyze whether the reasons are reasonable. If not, banks can thus modify the model.

%Authors are required to include a statement of the broader impact of their work, including its ethical aspects and future societal consequences. 
%Authors should discuss both positive and negative outcomes, if any. For instance, authors should discuss a) 
%who may benefit from this research, b) who may be put at disadvantage from this research, c) what are the consequences of failure of the system, and d) whether the task/method leverages
%biases in the data. If authors believe this is not applicable to them, authors can simply state this.

%Use unnumbered first level headings for this section, which should go at the end of the paper. {\bf Note that this section does not count towards the eight pages of content that are allowed.}

\bibliographystyle{plain}

\begin{thebibliography}{10}

\bibitem{belkin2006manifold}
Mikhail Belkin, Partha Niyogi, and Vikas Sindhwani.
\newblock Manifold regularization: A geometric framework for learning from
    labeled and unlabeled examples.
\newblock {\em Journal of machine learning research}, 2006.

\bibitem{chen2019measuring}
Deli Chen, Yankai Lin, Wei Li, Peng Li, Jie Zhou, and Xu~Sun.
\newblock Measuring and relieving the over-smoothing problem for graph neural
    networks from the topological view.
\newblock {\em AAAI Conference on Artificial Intelligence (AAAI)}, 2020.

\bibitem{DBLP:conf/nips/DefferrardBV16}
Micha{\"{e}}l Defferrard, Xavier Bresson, and Pierre Vandergheynst.
\newblock Convolutional neural networks on graphs with fast localized spectral
    filtering.
\newblock In {\em Advances in Neural Information Processing Systems (NIPS)},
    2016.

\bibitem{NIPS2017_6703}
Will Hamilton, Zhitao Ying, and Jure Leskovec.
\newblock Inductive representation learning on large graphs.
\newblock In {\em Advances in Neural Information Processing Systems, (NIPS)},
    2017.

\bibitem{DBLP:conf/cvpr/HeZRS16}
Kaiming He, Xiangyu Zhang, Shaoqing Ren, and Jian Sun.
\newblock Deep residual learning for image recognition.
\newblock In {\em Conference on Computer Vision and Pattern
    Recognition,{CVPR}}, 2016.

\bibitem{kipf2017gcn}
Thomas~N. Kipf and Max Welling.
\newblock Semi-supervised classification with graph convolutional networks.
\newblock In {\em International Conference on Learning Representations,
    (ICLR)}, 2017.

\bibitem{klicpera_predict_2019}
Johannes Klicpera, Aleksandar Bojchevski, and Stephan G{\"u}nnemann.
\newblock Predict then propagate: Graph neural networks meet personalized
    pagerank.
\newblock In {\em International Conference on Learning Representations (ICLR)},
    2019.

\bibitem{DBLP:conf/aaai/LiHW18}
Qimai Li, Zhichao Han, and Xiao{-}Ming Wu.
\newblock Deeper insights into graph convolutional networks for semi-supervised
    learning.
\newblock In {\em AAAI Conference on Artificial Intelligence, (AAAI)}, 2018.

\bibitem{Ma2019}
Jianxin Ma, Peng Cui, Kun Kuang, Xin Wang, and Wenwu Zhu.
\newblock {Disentangled Graph Convolutional Networks}.
\newblock {\em International Conference on Machine Learning, {ICML}}, 2019.

\bibitem{mccallum2000automating}
Andrew~Kachites McCallum, Kamal Nigam, Jason Rennie, and Kristie Seymore.
\newblock Automating the construction of internet portals with machine
    learning.
\newblock {\em Information Retrieval}, 2000.

\bibitem{namata2012query}
Galileo Namata, Ben London, Lise Getoor, Bert Huang, and UMD EDU.
\newblock Query-driven active surveying for collective classification.
\newblock In {\em International Workshop on Mining and Learning with Graphs},
    2012.

\bibitem{Perozzi2014}
Bryan Perozzi, Rami Al-Rfou, and Steven Skiena.
\newblock {DeepWalk: Online learning of social representations}.
\newblock {\em the ACM SIGKDD International Conference on Knowledge Discovery
    and Data Mining, (KDD)}, 2014.

\bibitem{sen2008collective}
Prithviraj Sen, Galileo Namata, Mustafa Bilgic, Lise Getoor, Brian Galligher,
    and Tina Eliassi-Rad.
\newblock Collective classification in network data.
\newblock {\em AI magazine}, 2008.

\bibitem{shchur2018pitfalls}
Oleksandr Shchur, Maximilian Mumme, Aleksandar Bojchevski, and Stephan
    G{\"u}nnemann.
\newblock Pitfalls of graph neural network evaluation.
\newblock In {\em Relational Representation Learning Workshop (R2L)}, 2018.

\bibitem{talukdar2009new}
Partha~Pratim Talukdar and Koby Crammer.
\newblock New regularized algorithms for transductive learning.
\newblock In {\em Joint European Conference on Machine Learning and Knowledge
    Discovery in Databases}, 2009.

\bibitem{triguero2015self}
Isaac Triguero, Salvador Garc{\'\i}a, and Francisco Herrera.
\newblock Self-labeled techniques for semi-supervised learning: taxonomy,
    software and empirical study.
\newblock {\em Knowledge and Information systems}, 2015.

\bibitem{DBLP:conf/iclr/VelickovicCCRLB18}
Petar Velickovic, Guillem Cucurull, Arantxa Casanova, Adriana Romero, Pietro
    Li{\`{o}}, and Yoshua Bengio.
\newblock Graph attention networks.
\newblock In {\em International Conference on Learning Representations,
    {ICLR}}, 2018.

\bibitem{DBLP:conf/icml/WuSZFYW19}
Felix Wu, Amauri H.~Souza Jr., Tianyi Zhang, Christopher Fifty, Tao Yu, and
    Kilian~Q. Weinberger.
\newblock Simplifying graph convolutional networks.
\newblock In {\em International Conference on Machine Learning,{ICML}}, 2019.

\bibitem{DBLP:conf/iclr/XuHLJ19}
Keyulu Xu, Weihua Hu, Jure Leskovec, and Stefanie Jegelka.
\newblock How powerful are graph neural networks?
\newblock In {\em International Conference on Learning Representations,
    {ICLR}}, 2019.

\bibitem{DBLP:conf/icml/YangCS16}
Zhilin Yang, William~W. Cohen, and Ruslan Salakhutdinov.
\newblock Revisiting semi-supervised learning with graph embeddings.
\newblock In {\em International Conference on Machine Learning,{ICML}}, 2016.

\bibitem{zhou2004learning}
Dengyong Zhou, Olivier Bousquet, Thomas~N Lal, Jason Weston, and Bernhard
    Sch{\"o}lkopf.
\newblock Learning with local and global consistency.
\newblock In {\em Advances in neural information processing systems, (NIPS)},
    2004.

\bibitem{zhu2003semi}
Xiaojin Zhu, Zoubin Ghahramani, and John~D Lafferty.
\newblock Semi-supervised learning using gaussian fields and harmonic
    functions.
\newblock In {\em the International conference on Machine learning (ICML)},
    2003.

\bibitem{zhu2005semi}
Xiaojin Zhu, John Lafferty, and Ronald Rosenfeld.
\newblock {\em Semi-supervised learning with graphs}.
\newblock PhD thesis, Carnegie Mellon University, language technologies
    institute, school of language technologies institute, 2005.

\end{thebibliography}

\clearpage
\appendix
\section{Hyper-parameters and data split}
\label{appendix:hyperparams}

\begin{table}[h]
\centering
\caption{Detailed hyper-parameters (hidden units/weight decay/learning rate/dropout rate/(batch size for MCGL-UM))}
    \begin{tabular}{cccc}
    \hline
    Dataset     & GCN             & GCN*             & MCGL-UM \\
    \hline
    CORA        & 32/0.0005/0.005/0.7 & 32/0.0005/0.01/0.7 & 32/0.001/0.005/0.5/50 \\
    CiteSeer    & 64/0.001/0.05/0.6 & 64/0.001/0.05/0.4 & 64/0.001/0.005/0.3/200 \\
    PubMed      & 32/0.0005/0.05/0.3 & 32/0.0005/0.005/0.5 & 32/0.001/0.005/0.5/50 \\
    MS Academic & 128/0.0005/0.01/0.6 & 128/0.0005/0.005/0.7 & 128/0.0001/0.005/0.5/200 \\
    \hline
    \end{tabular}
\label{hyperparams}
\end{table}

\begin{table}[h]
\centering
\caption{Original classification accuracy of all models on the four datasets with the specified hyper-parameters in Table \ref{hyperparams}.}
    \begin{tabular}{ccccc}
    \hline
    Dataset     & CORA          & CiteSeer      & PubMed        & MS Academic \\
    \hline
    GCN1        & $83.68\pm0.31$ & $72.32\pm0.35$ & $79.48\pm0.26$ & $92.32\pm0.22$ \\
    GCN2        & $83.24\pm0.31$ & $72.43\pm0.65$ & $79.41\pm0.40$ & $92.18\pm0.17$ \\
    MCGL-UM     & $83.53\pm0.56$ & $72.22\pm0.38$ & $77.91\pm0.70$ & $90.89\pm0.80$ \\
    \hline
    \end{tabular}
\label{original_acc}
\end{table}

The data split follows the split strategy of few-shot learning:
The training set consists of 20 random samples from each class.
There is no intersection between training set, validation set and test set. 
For CORA, Citeseer, and PubMed, we use the data division used by GCN~\cite{kipf2017gcn}. 
For MS Academic, we split data ourselves. 

Table \ref{hyperparams} lists the detailed hyper-parameters used for the four real-world datasets -- CORA, CiteSeer, PubMed and MS Academic, and all models -- traditional GCN, GCN* and MCGL-UM.
The strategy of deciding on hyperparameters is: vary one hyper-parameter per time while maintaining the others to find the optimal one; fix the hyper-parameters that have found the optimal value to the optimal, and continue to find the best value for the remaining. 
Fix the depth of the graph operation (GCO and MCGL) and the depth of the neural network to 2, adjusting the other super-parameters:
number of hidden units, learning rate, weight decay (L2-norm), dropout rate, and batch size.
GCN does not use batch-training strategy, so it has no batch size. 
The search spaces for hyper-parameters are as follows:
\begin{itemize}
    \item Number of hidden units: \{32, 64, 128\};
    \item Weight decay of Adam optimizer: \{0.005, 0.001, 0.0005, 0.0001\};
    \item Learning rate of Adam optimizer: \{0.05, 0.01, 0.005, 0.001\};
    \item Dropout rate: \{0.3, 0.4, 0.5, 0.6, 0.7\};
    \item \# Batch size: \{50, 100, 200, 300, 500\};
\end{itemize}
After finding the optimal hyper-parameters, they are fixed to the optimal value when doing other experiments (\eg graph noise rate and depth GCO experiments).
Table \ref{original_acc} lists the original classification accuracy of all models on the four datasets using the hyper-parameters listed above.

\section{Detailed experimental results}
\label{appendix:acc}

Table \ref{acc_noise_MCGL} and \ref{acc_noise_GCN} list detailed experimental results with mean value and sample standard deviation of comparing MCGL-UM and GCN models with reduced noise rate under fixed depth. The depth of MCGL-UM and GCN model are both set to 2.

Table \ref{acc_depth_noise_origin}-\ref{acc_depth_noise_0} list detailed experimental results with mean value and sample standard deviation of the classification accuracy of GCN* model with reduced noise rate under different depth.

We still use the four real-world datasets -- CORA, CiteSeer, PubMed and MS Academic. In the experiments, we manually reduce the noise rate to different levels by randomly eliminating bad edges in the adjacency matrix.

\begin{table}[t]
\centering
\caption{Percentile classification accuracy of MCGL-UM with different reduced noise rate.}
\begin{tabular}{ccccc}
\hline
Dataset & CORA & CiteSeer & PubMed & MS Academic \\
\hline
MCGL-UM model \\
\hline
$noise\ rate=18\%$ & $83.24\pm0.40$ & $73.68\pm0.69$ & $79.59\pm0.50$ & $92.49\pm0.33$ \\
$noise\ rate=15\%$ & $83.68\pm0.50$ & $73.70\pm1.29$ & $80.37\pm0.60$ & $93.34\pm0.29$ \\
$noise\ rate=12\%$ & $84.29\pm0.42$ & $74.00\pm1.27$ & $81.23\pm0.46$ & $94.26\pm0.24$ \\
$noise\ rate=9\%$ & $85.24\pm0.41$ & $74.72\pm0.91$ & $82.05\pm0.55$ & $95.23\pm0.22$ \\
$noise\ rate=6\%$ & $86.67\pm0.64$ & $74.68\pm0.71$ & $83.04\pm0.50$ & $96.23\pm0.16$ \\
$noise\ rate=3\%$ & $87.74\pm0.42$ & $75.01\pm0.69$ & $84.09\pm0.59$ & $97.20\pm0.15$ \\
$noise\ rate=0\%$ & $89.43\pm0.32$ & $76.60\pm1.09$ & $85.17\pm0.38$ & $98.15\pm0.08$ \\
\hline
\end{tabular}
\label{acc_noise_MCGL}
\end{table}

\begin{table}[t]
\centering
\caption{Percentile classification accuracy of GCN with different reduced noise rate.}
\begin{tabular}{ccccc}
\hline
Dataset & CORA & CiteSeer & PubMed & MS Academic \\
\hline
GCN model \\
\hline
$noise rate=18\%$ & $83.59\pm0.90$ & $72.89\pm0.78$ & $78.86\pm0.60$ & $91.30\pm0.65$ \\
$noise rate=15\%$ & $83.63\pm0.97$ & $73.26\pm0.86$ & $79.46\pm0.51$ & $92.39\pm0.56$ \\
$noise rate=12\%$ & $84.86\pm0.80$ & $73.67\pm0.71$ & $80.27\pm0.85$ & $93.59\pm0.46$ \\
$noise rate=9\%$ & $86.04\pm0.86$ & $74.30\pm0.89$ & $81.34\pm0.57$ & $94.94\pm0.27$ \\
$noise rate=6\%$ & $87.34\pm0.69$ & $75.05\pm0.81$ & $82.19\pm0.59$ & $96.09\pm0.34$ \\
$noise rate=3\%$ & $88.54\pm0.99$ & $75.82\pm0.81$ & $83.53\pm0.61$ & $97.41\pm0.15$ \\
$noise rate=0\%$ & $90.55\pm0.39$ & $77.57\pm0.65$ & $85.18\pm0.25$ & $98.79\pm0.08$ \\
\hline
\end{tabular}
\label{acc_noise_GCN}
\end{table}

\begin{table}[t]
\centering
\caption{Percentile classification accuracy with increasing depth ($K$) under original noise rate.}
\begin{tabular}{ccccc}
\hline
Dataset & CORA & CiteSeer & PubMed & MS Academic \\
\hline
$noise\ rate$ & 19.00\% & 26.45\% & 19.76\% & 19.19\% \\
\hline
$K=0$ & $57.79\pm0.80$ & $58.52\pm1.64$ & $73.33\pm0.56$ & $90.22\pm0.39$ \\
$K=1$ & $80.27\pm0.45$ & $69.79\pm0.69$ & $77.59\pm0.36$ & $92.29\pm0.34$ \\
$K=2$ & $83.54\pm0.32$ & $72.44\pm0.50$ & $79.49\pm0.17$ & $92.26\pm0.29$ \\
$K=3$ & $83.87\pm0.42$ & $72.35\pm0.63$ & $78.76\pm0.20$ & $91.47\pm0.29$ \\
$K=4$ & $83.73\pm0.75$ & $72.22\pm0.74$ & $79.29\pm0.21$ & $90.80\pm0.33$ \\
$K=5$ & $83.22\pm0.51$ & $72.29\pm0.63$ & $79.54\pm0.20$ & $90.15\pm0.37$ \\
$K=6$ & $82.77\pm0.62$ & $72.42\pm0.95$ & $79.38\pm0.33$ & $89.51\pm0.42$ \\
$K=7$ & $82.70\pm0.64$ & $72.50\pm0.45$ & $79.58\pm0.14$ & $89.00\pm0.45$ \\
$K=8$ & $82.46\pm0.54$ & $72.36\pm0.54$ & $79.78\pm0.23$ & $88.25\pm0.48$ \\
$K=9$ & $82.31\pm0.28$ & $72.55\pm0.54$ & $79.82\pm0.13$ & $87.60\pm0.65$ \\
$K=10$ & $82.17\pm0.28$ & $72.31\pm0.57$ & $79.86\pm0.25$ & $86.99\pm0.72$ \\
$K=11$ & $82.01\pm0.63$ & $71.96\pm0.45$ & $79.76\pm0.21$ & $86.36\pm0.81$ \\
$K=12$ & $81.86\pm0.55$ & $72.05\pm0.68$ & $79.66\pm0.23$ & $85.62\pm0.91$ \\
$K=13$ & $81.80\pm0.62$ & $72.02\pm0.58$ & $79.68\pm0.30$ & $85.10\pm1.12$ \\
$K=14$ & $81.49\pm0.65$ & $72.01\pm0.56$ & $79.66\pm0.27$ & $84.28\pm1.25$ \\
$K=15$ & $81.61\pm0.49$ & $72.02\pm0.71$ & $79.39\pm0.37$ & $83.44\pm1.69$ \\
\hline
\end{tabular}
\label{acc_depth_noise_origin}
\end{table}

\begin{table}[t]
\centering
\caption{Percentile classification accuracy with increasing depth ($K$) under noise rate of 10\%.}
\begin{tabular}{ccccc}
\hline
Dataset & CORA & CiteSeer & PubMed & MS Academic \\
\hline
$noise\ rate$ & 10\% & 10\% & 10\% & 10\% \\
\hline
$K=0$ & $57.79\pm0.80$ & $58.52\pm1.64$ & $73.33\pm0.56$ & $90.22\pm0.39$ \\
$K=1$ & $82.57\pm0.54$ & $71.37\pm0.93$ & $78.36\pm0.62$ & $94.48\pm0.33$ \\
$K=2$ & $85.40\pm0.67$ & $74.86\pm1.04$ & $81.77\pm0.49$ & $95.04\pm0.21$ \\
$K=3$ & $85.72\pm0.88$ & $74.86\pm1.31$ & $81.99\pm0.54$ & $94.69\pm0.21$ \\
$K=4$ & $85.97\pm0.98$ & $74.80\pm1.06$ & $82.69\pm0.37$ & $94.38\pm0.21$ \\
$K=5$ & $85.76\pm0.59$ & $75.01\pm0.84$ & $83.15\pm0.62$ & $94.13\pm0.21$ \\
$K=6$ & $85.58\pm0.71$ & $75.08\pm1.14$ & $83.88\pm0.55$ & $93.88\pm0.19$ \\
$K=7$ & $85.30\pm0.75$ & $75.12\pm0.83$ & $83.86\pm0.43$ & $93.64\pm0.15$ \\
$K=8$ & $85.27\pm0.84$ & $75.27\pm0.92$ & $84.05\pm0.38$ & $93.33\pm0.20$ \\
$K=9$ & $85.29\pm0.79$ & $75.15\pm1.24$ & $84.02\pm0.56$ & $93.03\pm0.18$ \\
$K=10$ & $85.28\pm0.84$ & $75.16\pm1.34$ & $83.98\pm0.47$ & $92.75\pm0.27$ \\
$K=11$ & $85.07\pm0.93$ & $75.18\pm0.91$ & $83.94\pm0.67$ & $92.49\pm0.34$ \\
$K=12$ & $84.94\pm0.83$ & $75.19\pm1.00$ & $83.74\pm0.73$ & $91.96\pm0.89$ \\
$K=13$ & $84.70\pm0.88$ & $75.10\pm1.22$ & $83.76\pm0.70$ & $91.72\pm0.91$ \\
$K=14$ & $84.67\pm0.93$ & $75.12\pm1.24$ & $83.45\pm0.57$ & $91.40\pm1.04$ \\
$K=15$ & $84.62\pm0.90$ & $75.15\pm0.75$ & $83.45\pm0.47$ & $91.03\pm1.20$ \\
\hline
\end{tabular}
\label{acc_depth_noise_0.1}
\end{table}

\begin{table}[t]
\centering
\caption{Percentile classification accuracy with increasing depth ($K$) under noise rate of 0.}
\begin{tabular}{ccccc}
\hline
Dataset & CORA & CiteSeer & PubMed & MS Academic \\
\hline
$noise\ rate$ & 0 & 0 & 0 & 0 \\
\hline
$K=0$ & $57.79\pm0.80$ & $58.52\pm1.64$ & $73.33\pm0.56$ & $90.22\pm0.39$ \\
$K=1$ & $86.09\pm0.54$ & $73.97\pm0.71$ & $80.27\pm0.38$ & $97.01\pm0.22$ \\
$K=2$ & $89.66\pm0.30$ & $77.08\pm0.39$ & $85.01\pm0.26$ & $98.21\pm0.07$ \\
$K=3$ & $90.69\pm0.30$ & $77.94\pm0.71$ & $86.48\pm0.37$ & $98.61\pm0.07$ \\
$K=4$ & $91.23\pm0.25$ & $78.69\pm0.73$ & $87.94\pm0.31$ & $98.79\pm0.08$ \\
$K=5$ & $91.47\pm0.25$ & $79.33\pm0.98$ & $88.51\pm0.30$ & $98.86\pm0.07$ \\
$K=6$ & $91.63\pm0.32$ & $79.44\pm0.77$ & $89.18\pm0.37$ & $98.92\pm0.07$ \\
$K=7$ & $91.81\pm0.37$ & $80.01\pm0.95$ & $89.34\pm0.49$ & $98.98\pm0.07$ \\
$K=8$ & $91.79\pm0.38$ & $79.84\pm0.69$ & $89.38\pm0.40$ & $98.98\pm0.08$ \\
$K=9$ & $91.90\pm0.34$ & $80.34\pm0.98$ & $89.57\pm0.32$ & $99.01\pm0.08$ \\
$K=10$ & $91.81\pm0.30$ & $80.17\pm0.90$ & $89.90\pm0.30$ & $99.02\pm0.06$ \\
$K=11$ & $92.02\pm0.46$ & $80.23\pm0.87$ & $90.37\pm0.37$ & $99.02\pm0.07$ \\
$K=12$ & $92.16\pm0.41$ & $80.43\pm0.68$ & $90.78\pm0.21$ & $99.03\pm0.07$ \\
$K=13$ & $92.14\pm0.53$ & $80.68\pm0.75$ & $91.10\pm0.23$ & $99.04\pm0.06$ \\
$K=14$ & $92.06\pm0.46$ & $80.63\pm0.82$ & $91.21\pm0.35$ & $99.04\pm0.07$ \\
$K=15$ & $92.14\pm0.42$ & $80.66\pm0.65$ & $91.44\pm0.20$ & $99.04\pm0.08$ \\
\hline
\end{tabular}
\label{acc_depth_noise_0}
\end{table}

\end{document}